\pdfoutput=1

\documentclass[11pt]{article}

\usepackage[]{acl}

\usepackage{times}
\usepackage{latexsym}
\usepackage{amsmath}
\usepackage{graphicx} 
\usepackage{pifont}
\usepackage{multirow}
\usepackage{enumitem}
\usepackage{booktabs} 

\usepackage[T1]{fontenc}

\usepackage[utf8]{inputenc}

\usepackage{microtype}

\usepackage{inconsolata}

\title{Superfiltering: Weak-to-Strong Data Filtering for Fast Instruction-Tuning}

\author{ 
  \textbf{Ming Li}\textsuperscript{1}, 
  \textbf{Yong Zhang}\textsuperscript{2}, 
  \textbf{Shwai He}\textsuperscript{1}, 
  \textbf{Zhitao Li}\textsuperscript{2}, 
  \textbf{Hongyu Zhao}\textsuperscript{1}, \\
    \textbf{Jianzong Wang}\textsuperscript{2}, 
  \textbf{Ning Cheng}\textsuperscript{2}, 
  \textbf{Tianyi Zhou}\textsuperscript{1}\thanks{Corresponding author} \\
  \textsuperscript{1}University of Maryland \textsuperscript{2}Ping An Technology (Shenzhen) Co., Ltd.\\
  minglii@umd.edu, jzwang@188.com, tianyi@umd.edu \\
  Project: \url{https://github.com/tianyi-lab/Superfiltering}
}

\begin{document}
    \maketitle
\begin{abstract}
Instruction tuning is critical to improve LLMs but usually suffers from low-quality and redundant data. 
Data filtering for instruction tuning has proved important in improving both the efficiency and performance of the tuning process. 
But it also leads to extra cost and computation due to the involvement of LLMs in this process. 
To reduce the filtering cost, we study {\bf Superfiltering}: {\it Can we use a smaller and weaker model to select data for finetuning a larger and stronger model?} 
Despite the performance gap between weak and strong language models, we find their highly consistent capability to perceive instruction difficulty and data selection results. 
This enables us to use a much smaller and more efficient model to filter the instruction data used to train a larger language model. Not only does it largely speed up the data filtering, but the filtered-data-finetuned LLM achieves even better performance on standard benchmarks. 
Extensive experiments validate the efficacy and efficiency of our approach. 

\end{abstract}

\begin{figure}[!t]
\centering 
\includegraphics[width=0.45\textwidth]{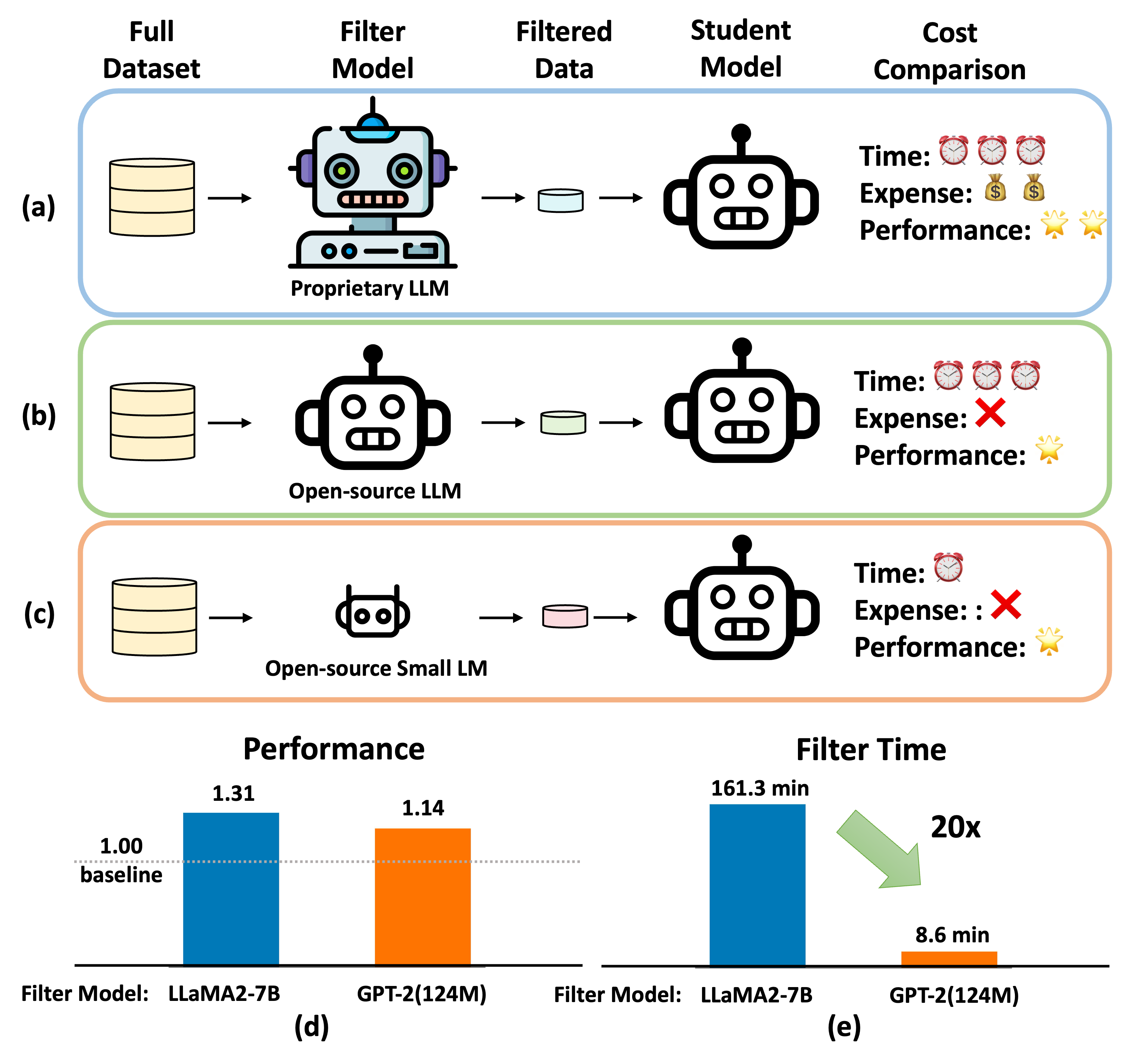} 
\caption{
\textbf{Top:} Comparison of data filtering for instruction tuning of a student model. \textbf{(a)} The filter model is a strong proprietary LLM, e.g. ChatGPT, which can be time-consuming and expensive but usually performs promisingly. \textbf{(b)} The filter model is the student model itself or a similar-sized open-source LLM, which is still time-consuming but free to use. \textbf{(c)} \textbf{Weak-to-strong superfiltering} proposed by this paper, which utilizes a much smaller filter model, e.g. GPT-2, to train a stronger student LLM. We find it costs much less time but maintains the performance. 
\textbf{Bottom:} Comparisons of two student models finetuned using 5\% data selected by LLaMA2-7B and GPT-2 from the Alpaca dataset. \textbf{(d)} Both models trained on 5\% data outperform the baseline model trained on 100\% data. \textbf{(e)} GPT-2 as the superfilter speeds up data filtering by $\sim20$ times. 
} 
\label{diff_models} 
\end{figure}

\section{Introduction}



Earlier works of instruction tuning on Large Language Models (LLMs) \cite{NEURIPS2020_1457c0d6, openai2023gpt4, touvron2023llama, jiang2023mistral} focus on creating large, varied, and high-quality datasets of various tasks with responses curated by human experts~\cite{khashabi-etal-2020-unifiedqa, ye-etal-2021-crossfit, wei2022finetuned, wang-etal-2022-super, du-etal-2022-glm}, which can be bottlenecked by the intensive human labor. 
An alternative is to generate the data by a powerful teacher LLM~\cite{wang-etal-2023-self-instruct, alpaca, xu2023wizardlm, Li2023ReflectionTuningDR, Li2024SelectiveRS, Li2024CanLS, xu2024survey} but the quality is hard to control and largely depends on the teacher. 
LIMA~\cite{zhou2023lima} finds that a mere 1,000 human-crafted high-quality data could significantly improve an LLM's instruction-following capability, based on which they posit that LLMs acquire most knowledge during the pretraining and thus a few data suffices for instruction tuning.

To further free the human labor in data curation and accelerate the instruction tuning process, a line of recent works apply an extra filter algorithm to select data from the existing dataset. However, the model used in the filtering process usually needs to be as powerful as ChatGPT~\cite{chen2023alpagasus, lu2023instag} or requires additional reward data training \cite{du2023mods, bukharin2023data}, or is the student model itself~\cite{cherry, Li2024SelectiveRS}, which leads to additional expensive cost and latency due to the inference on these large filter models, especially when the original dataset is large while only a tiny fraction of data needs to be selected. These paradigms are presented in Figure \ref{diff_models} (a) and (b). To reduce the filtering cost, we study \textbf{Superfiltering}: \textbf{Can we use a smaller and weaker model as a filter to select instruction-tuning data for training a larger and stronger model?} This was first studied for training small classification models by \citet{Coleman2020Selection} while the effectiveness on the open-domain instruction dataset is un-explored. Recently, Weak-to-Strong Generalization \cite{burns2023weaktostrong} proposes to utilize a weaker ChatGPT to generate data used to finetune a stronger GPT4 model, which shares a similar spirit with our Superfiltering as depicted in Figure \ref{diff_models}(c).

In Superfitering, we find that a smaller and weaker GPT-2 (124M) \cite{radford2019language} suffices to replace previously used large filter models and select high-quality instruction tuning data used to finetune a much larger LLaMA2 (7B or 13B). This is motivated by our main discovery of filter models' consistency on two data statistical metrics, perplexity and instruction-following difficulty (IFD) score~\cite{cherry}. Despite the differences in scales across different filter models, their rankings of the same instruction tuning dataset are surprisingly consistent, as demonstrated by the large rank correlation coefficients evaluated on different models and datasets. Our thorough empirical study implies that weaker language models possess a capability consistent with their stronger counterparts in comprehending and discerning the difficulty of diverse instructions, though they may differ in other skills like reasoning and generalization. 

In extensive experiments, our Superfiltering strategy using GPT-2 as the filter, as exemplified on several widely used instruction datasets, brings significant speedups to data filtering for instruction tuning. By utilizing only 5\% of the original data volume, Superfilter allows us to attain LLMs comparable, and in some instances superior, to those achieved by training with full data. Our main contributions can be summarized in three folds:
\begin{itemize}[leftmargin=*]
\item \textbf{Weak-to-Strong Consistency on Data Filtering}: We reveal the strong consistency between small and large LLMs in perceiving and evaluating the difficulty of instruction tuning data. \looseness-1 
\item \textbf{Efficent Superfiltering Strategy}: We propose the first method of Superfiltering that utilizes a small LM, e.g., GPT-2 (124M), to select data for instruction tuning, and brings significant speedups to LLM finetuning pipeline.  
\item \textbf{Efficacy of Selected Training Data}: Superfiltering is precise in allocating high-quality and informative data improving LLM instruction tuning.  
\vspace{-2mm}
\end{itemize}

\begin{figure}[!t]
\centering 
\includegraphics[width=0.50\textwidth]{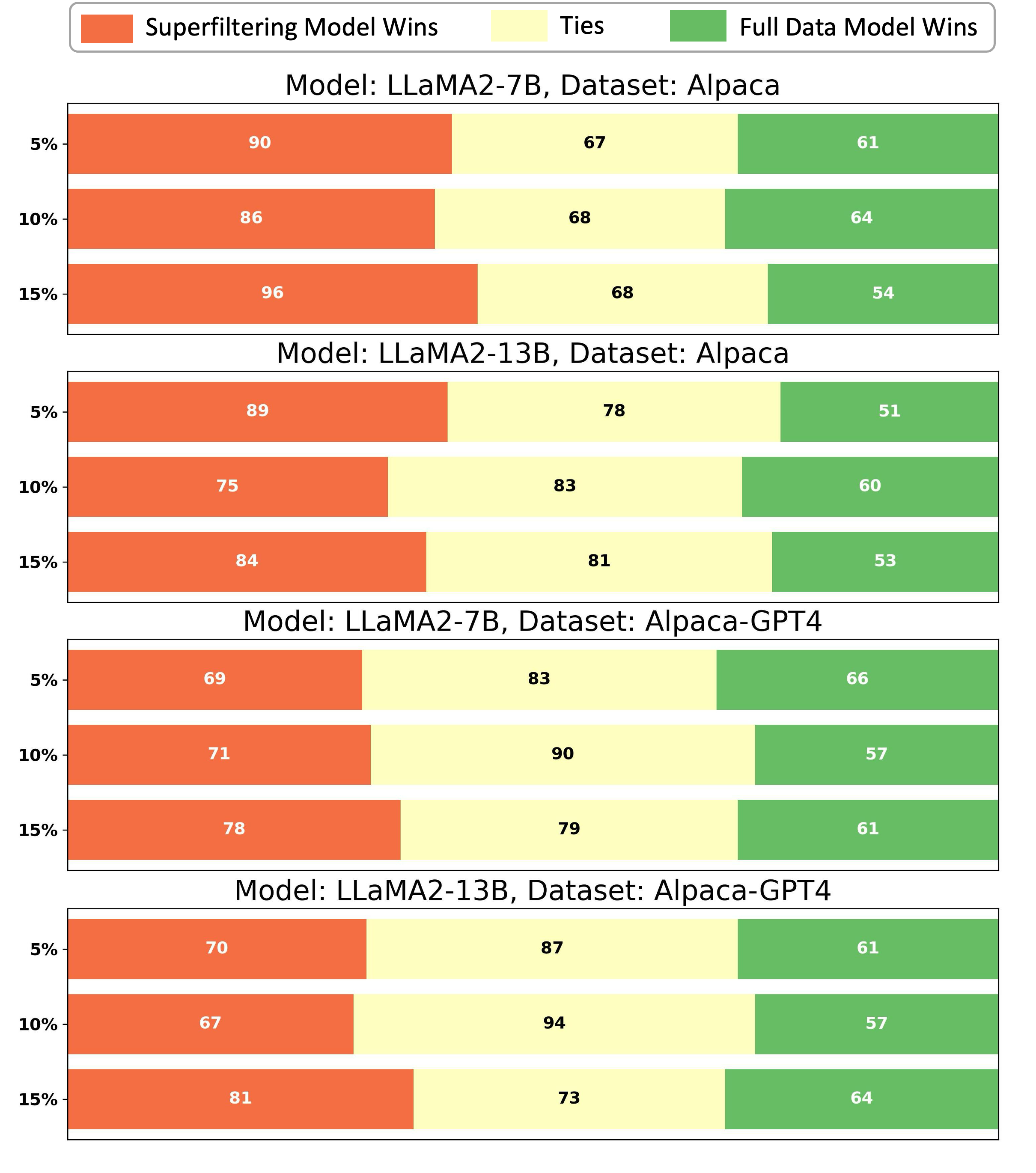} 
\caption{
Pairwise comparison between each model finetuned using Superfiltered data (5\%, 10\%, and 15\% of the original dataset) and the full-data (100\%) finetuned model. We report results for two base models (LLaMA2-7B/13B) and two datasets (Alpaca and Alpaca-GPT4 datasets). The win-tie-lose is judged by GPT-4 given two models' responses to each instruction from WizardLM test set. \looseness-1
} 
\label{pair} 
\end{figure}

\section{Problem Formulation}

\subsection{Preliminaries}

We define a dataset as $D$, containing  $n$ triplets $x = (Instruction, [Input], Response)$ as the instruction tuning data samples. Earlier instructing tuning samples mostly contain separated $instruction$ and $input$ segments of better controls \cite{wang-etal-2022-super, longpre2023flan, alpaca}, while most of the current datasets directly merge the inputs to instructions \cite{zhou2023lima, vicuna2023, xu2023wizardlm, Li2023ReflectionTuningDR}. For simplicity, we define $x = map(Instruction, [Input]) $ as the complete instruction and $y$ as the corresponding response. The mapping function could be the simple concatenation with some control tokens. Thus $D=\{(x_1, y_1), (x_2, y_2), \ldots, (x_n, y_n)\}$ represents a collection of $n$ instruction-response pairs. 

\paragraph{Perplexity}

In the instruction tuning setting, the model is trained to maximize the likelihood of response given the corresponding instruction as the condition. Hence, perplexity can be a potential metric to measure the difficulty. Specifically, the perplexity of a given sample $(x_i,y_i)$ is defined as: 

{\small
\begin{equation}
\text{PPL}(y_i|x_i) = \exp\left(
-\frac{1}{N}\sum_{j=1}^{N}\log p(y_{i,j}|x_i, y_{i,1}, ..., y_{i,j-1})\right) 
\end{equation}
}
where $N$ is the length of response $y_i$ and $y_{i,j}$ represents the $j$th token in the response $y_i$. 

\paragraph{IFD score}

\citet{cherry} firstly proposes a self-guided method in which no extra models are utilized but needs to calculate Instruction-Following Difficulty (IFD) scores based on the pre-experienced LLM or original pre-trained LLM. 
The IFD score is a pure statistical metric, that compares the losses or perplexities when the model generates a response $y_i$ with and without instructional context $x_1$, measuring how much help the instruction provides to the generation of the corresponding response. A higher IFD score, indicating less instructional help, suggests a greater difficulty. On the contrary, the low IFD score represents that the given instruction can directly benefit the language model largely even without further training, representing the easiness and necessity of the instruction. 
For a given instruction-following data pair, the IFD score is calculated as follows: 
\begin{equation}
    \text{IFD}(y_i|x_i) = \frac{\text{PPL}(y_i|x_i)}{\text{PPL}(y_i)}
\end{equation}
where \(\text{PPL}(y_i|x_i)\) and \(\text{PPL}(y_i)\) denote the perplexities of the given model in fitting response \(y_i\) with and without the instruction \(x_i\), respectively.

\subsection{Formulation and Motivations}

Superfiltering aims to find a data filtering score (1) that excels in identifying high-quality and informative training data, and (2) computed by a small and low-cost filter model without further training. To this end, we try to find a data evaluation metric consistent between weak and strong language models. \looseness-1

Given a candidate score, we investigate whether it is possible to utilize a much weaker language model, e.g. GPT-2, to calculate for the relatively stronger student model. We hypothesize that, although the intrinsic abilities between weak and strong language models vary dramatically, indicated by the discrepancies of perplexities on the pretraining stage, their ability to perceive instruction difficulty could be similar. To verify our hypothesis, experiments are conducted and presented in Section~\ref{sec:consistency}. 

To verify the hypothesis, we conduct a thorough empirical study of the consistency of perplexities computed by different language models on the same instruction-tuning dataset. In Section~\ref{sec:perplexity}, we focus on verifying the consistency of perplexity across weak-to-strong models by comparing their scale and orderings of samples on each dataset. The results show that though the scales vary drastically, the orderings remain consistent, which verifies our hypothesis. In Section~\ref{sec:IFD}, we conduct the same study on IFD scores, on which both the scales and the orderings are consistent across weak-to-strong models, indicating IFD score as a more promising score for Superfiltering than perplexity. 

Figure~\ref{pair} compares each Superfiltering-selected-data finetuned model and the full-data finetuned model by using GPT-4 as a judge to decide their numbers of wins/ties/losses on a test set of instructions. More details of the evaluation metric can be found in Section \ref{sec:eva}. Superfiltering-trained models always outperform the baseline given different base models, datasets, and selection ratios, demonstrating the effectiveness of our proposed weak-to-strong Superfiltering scheme.

\begin{table}[!tbh]
\centering
\small
\scalebox{0.7}{
\begin{tabular}{l|l|cc|ccc}
\toprule
\multirow{2}{*}{\textbf{Dataset}} & \multicolumn{1}{c|}{\textbf{Model}} & \multicolumn{2}{c|}{\textbf{Rank Correlation} $\uparrow$} & \multicolumn{3}{|c}{\textbf{Overlap Ratios} $\uparrow$} \\
& Name & Perplexity  & IFD score & 5\% & 10\% & 15\% \\
\midrule
\multirow{5}{*}{Alpaca} 
 & GPT-2        & 0.726 & 0.679 & 0.28 & 0.41 & 0.49  \\
 & GPT-2-large  & 0.790 & 0.682 & 0.26 & 0.40 & 0.50  \\
 & GPT-2-XL    & 0.802 & 0.693 & 0.27 & 0.40 & 0.49  \\
 & GPT-NEO       & 0.846 & 0.802 & 0.38 & 0.51 & 0.59  \\
 & LLaMA2-7B     & 1.000 & 1.000 & 1.00 & 1.00 & 1.00  \\
\midrule
\multirow{5}{*}{Alpaca-GPT4} 
 & GPT-2      & 0.730 & 0.788 & 0.24 & 0.40 & 0.51  \\
 & GPT-2-large  & 0.795 & 0.820 & 0.21 & 0.36 & 0.48  \\
 & GPT-2-XL    & 0.800 & 0.818 & 0.18 & 0.33 & 0.45  \\
 & GPT-NEO       & 0.842 & 0.876 & 0.33 & 0.52 & 0.62  \\ 
 & LLaMA2-7B    & 1.000 & 1.000 & 1.00 & 1.00 & 1.00  \\
\midrule
\multirow{5}{*}{Wizard 70k}  
 & GPT-2        & 0.763 & 0.802 & 0.42 & 0.54 & 0.61  \\
 & GPT-2-large  & 0.809 & 0.848 & 0.44 & 0.58 & 0.65  \\
 & GPT-2-XL    & 0.821 & 0.855 & 0.44 & 0.57 & 0.65  \\
 & GPT-NEO    & 0.857 & 0.893 & 0.52 & 0.63 & 0.69  \\ 
 & LLaMA2-7B    & 1.000 & 1.000 & 1.00 & 1.00 & 1.00  \\
\bottomrule
\end{tabular}
}
\caption{
The rank correlation coefficient (Spearman's $\rho$) and overlap ratio (of selected data) between LLaMA2-7B and smaller language models when applied as filter models on three widely-used instruction tuning datasets. When calculating Spearman's $\rho$, the samples are sorted by perplexity or IFD scores calculated by different filter models. For the overlap ratio, we consider three data filtering budgets, i.e., when 5\%, 10\%, or 15\% of the dataset are selected. The large rank coefficient between LLaMA2-7B and other smaller models indicates the consistency of different models in perceiving the difficulties of instruction tuning data. 
}
\label{tbl:trans}
\end{table}

\section{Weak-to-Strong Consistency}
\label{sec:consistency}

In this section, we delve into the hypothesis that weak and strong language models share a relatively consistent capability in perceiving the difficulties of instruction tuning samples.

\subsection{Weak-to-Strong Perplexity Consistency}
\label{sec:perplexity}

\begin{figure*}[t]
\centering 
\includegraphics[width=1.0\textwidth]{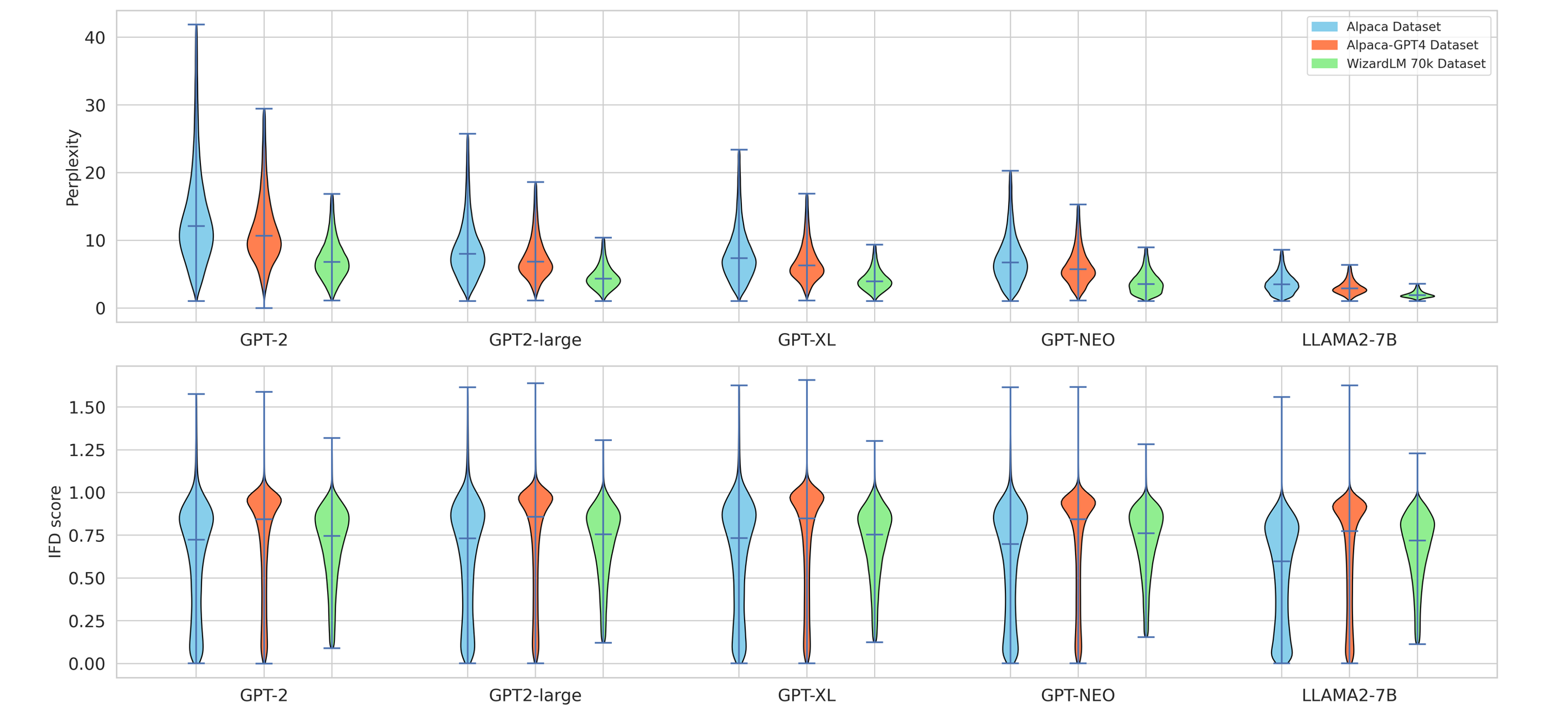} 
\caption{
The distributions of perplexity (top) and IFD score (bottom) computed by five models (left-to-right: weak-to-strong) on three instruction tuning datasets. \textbf{Observations:} (1) The scale of perplexity varies drastically across different models, indicating their difference in generation capability; (2) The scale of IFD scores is consistent across models, indicating their consistency in measuring difficulties. 
} 
\label{ifd_dis} 
\end{figure*}

As mentioned in the previous section, we first need to have a grasp of to what extent language models of different sizes are consistent with each other in understanding instructions and generating corresponding responses. Thus we calculate the perplexity scores of several pretrained language models, including relatively small language models like GPT-2 (124M), GPT-2-large (774M), GPT-2-XL (1.5B) \cite{radford2019language}, GPT-NEO (1.3B) \cite{gpt-neo} and recent relatively strong LLaMA2-7b \cite{touvron2023llama2}, on several instruction-tuning dataset including the Alpaca \cite{alpaca}, Alpaca-GPT4 \cite{peng2023instruction}, and WizardLM 70k \cite{xu2023wizardlm}.
The results are shown in Figure \ref{ifd_dis} (upper), where each box presents a perplexity distribution of a given dataset and language model. 
A clear tendency can be found that the stronger the language models are, the lower the perplexities are, which is consistent with the common beliefs for LLM pretraining: the better a language model, the lower this perplexity. \looseness-1

The above experimental results only showcase the perplexity scales of different models and neglect the potential perplexity ordering/ranking of different data samples, which is much more vital for data filtering. Thus to evaluate the similarity in perplexity ordering on a given dataset between different models, Spearman's rank correlation coefficient (Spearman's $\rho$) is utilized. Spearman's $\rho$ is a non-parametric measure used to assess the strength and direction of the relationship between two variables that are ranked or ordinal in nature. 
For two lists containing the same elements but different ordering, this value measures the similarity of the ordering in the range of $-1$ to $1$. The closer the value is to $1$, the more consistent the ordering of these two lists. 

Specifically, within each dataset $D$, we sort the data samples based on the perplexity scores calculated by different models, resulting in several lists containing the same data but different orders, noted as $D_{\text{PPL, GPT-2}}$, $D_{\text{PPL, LLaMA2-7B}}$, etc. Since most of the fine-tuned experiments in our work are implemented on the LLaMA2-7B model, we set $D_{\text{PPL, LLaMA2-7B}}$ as our standard sorted list and calculate the Spearman's $\rho$ between the sorted lists of small language models and LLaMA2-7B:
\begin{equation}
\rho_{\text{PPL, GPT-2}} = g(D_{\text{PPL, GPT-2}},D_{\text{PPL, LLaMA2-7B}} )
\end{equation}
where $g$ is the function of calculating this coefficient. All the resulting values on different instruction tuning datasets and different models are presented in Table \ref{tbl:trans}, the Spearman’s Coefficient-Perplexity column. 

From the results, we can see even the lowest coefficient value is still greater than $0.7$, calculated between GPT-2(124M) and LLaMA2-7B, and the highest coefficient value is greater than $0.85$, calculated between GPT-NEO(1.3B) and LLaMA2-7B. The values presented in the table are reasonably high, indicating the consistent capability of different models in perceiving instructions. Moreover, there is also a clear tendency that the stronger the language models are, the higher the coefficient values. 
Comparing the perplexity distributions in Figure \ref{ifd_dis} (upper) and the coefficient values in Table \ref{tbl:trans}, a clear consistency can be revealed: 
Despite the large variance in the scales of perplexities generated by different language models, representing the intrinsic abilities of different language models, the high consistency in the perplexity ordering indicates the similarity of them to understand instructions. That is to say, for a given instruction tuning sample, if the weak language models find it hard to generate based on the corresponding instruction, the strong models might probably feel the same way even though their probability of generating this response is much larger, and vice versa. 

This phenomenon directly provides a glance at the weak-to-strong perplexity consistency, which serves as the basis for utilizing weak language models as the proxies for strong language models. 


\subsection{Weak-to-Strong IFD Consistency}
\label{sec:IFD}

Though a clear consistency in the perplexities of different language models is revealed by the above experiments, the perplexity does not directly represent the difficulty or quality of the instruction tuning sample and is thus not able to be used for the data section. Thus we further extend our findings to the Instruction-Following Difficulty (IFD) score proposed by Cherry LLM \cite{cherry}. It is used to select a subset of high-quality samples from the given instruction-tuning dataset to train an LLM with better performance. 

Similarly, we calculate the IFD scores on different instruction-tuning datasets with different language models and draw their distributions as shown in Figure \ref{ifd_dis} (lower). We observe that though the perplexity scales vary noticeably between models, the IFD scales remain similar, indicating its potential to be the general selection metric for different models. Furthermore, the IFD-based Spearman's $\rho$ are also presented in Table \ref{tbl:trans} Spearman’s Coefficient-IFD score column. Similar to the perplexity-based coefficient values, IFD-based values also remain high, indicating a strong consistency of IFD rankings calculated on different models. Such a consistency validates the scalability of weaker models in evaluating instruction difficulty, indicating their adeptness at identifying complex instructions akin to their stronger counterparts. Another interesting phenomenon is that the IFD-based coefficient values are greater than perplexity-based values on high-quality datasets, e.g. Alpaca-GPT4 and WizardLM 70k, indicating an even higher consistency in IFD scores for these datasets. 

To provide an even further apparent glance at this consistency, we calculate the overlap ratio when utilizing IFD scores to select the high-quality subset. The performances of the LLMs could be slightly estimated by the overlap ratio due to the previous success of this metric. As the percentage threshold increases from $5 \%$ to $15\%$, there is a significant and growing overlap in the samples identified by the weaker models and strong models like the LLaMA2-7B model. Although the overlap is not complete, it is substantial, this increasing overlap with higher thresholds reinforces our hypothesis, affirming a consistent and scalable capability in instruction evaluation across models of varying sizes.  

This weak-to-strong IFD consistency directly verifies our hypothesis that language models with different sizes possess similar capabilities in understanding the difficulty of the instructions, even though their intrinsic abilities are varied. It means that the difficult instruction tuning samples defined by the IFD scores are probably ``generally'' difficult no matter what language model is utilized for the calculation. This phenomenon directly makes it possible to utilize weak language models as the proxies for strong language models for calculating the IFD scores, and thus, to select data for instruction tuning.

\subsection{Superfiltering}

From the above section, we observe that the IFD score is a highly consistent metric when calculating based on different instruction-tuning datasets and varied-size language models. Thus we propose ``Superfiltering'', the first approach utilizing only small language models, i.e. GPT-2 (124M) \cite{radford2019language} to filter data for the instruction tuning of modern LLMs. Superfiltering uses smaller, less resource-intensive models (referred to as ``weak'' models) as effective substitutes for larger models (referred to as ``strong'' models) in the data evaluations. For the first time, making this process so efficient as to put it into practical usage. 
Specifically, following \citet{cherry}, for the given instruction-tuning dataset, the GPT-2 model is directly used to calculate the IFD score of each sample. Then the top $k$-percent samples with the highest IFD scores under $1$ are selected for faster instruction tuning. 

\section{Experimental Setup}

\subsection{Datasets}

The Alpaca dataset \cite{alpaca} is developed by Stanford University, comprises 52,000 instruction-following samples, and was created using the self-instruct paradigm \cite{wang-etal-2023-self-instruct}. This dataset was generated by leveraging OpenAI's text-davinci-003 model. 
The Alpaca dataset represents a classical dataset with moderate qualities, to further verify our method on the originally high-quality dataset, we also implement our method on the Alpaca-GPT4 dataset \cite{peng2023instruction}, which contains the responses generated by GPT4.

\subsection{Implementation Details}

We utilize the prompt and code base from Vicuna \cite{vicuna2023} and flash attention \cite{dao2022flashattention} while the overall training arguments are aligned with the common training configuration. The Adam optimizer \cite{kingma2017adam}, with a $2\times10^{-5}$ learning rate for the LLaMA2-7B model \cite{touvron2023llama2} and a $1\times10^{-5}$ learning rate for the LLaMA2-13B model, and a batch size of $128$, steer the training across three epochs with a max length of $2048$. The warmup rate is set to $0.03$. \looseness-1

\subsection{Evaluation Metrics}
\label{sec:eva}

\subsubsection{Automatic Evaluation}

To evaluate the effectiveness of our method, we utilize 3 commonly used automatic evaluation metrics, including (1) \textbf{Pair-wise Comparison}, (2) \textbf{Open LLM Leaderboard} and (3) \textbf{Alpaca Eval}. \footnote{Detailed description of evaluation metrics can be found in Appendix \ref{evaluation}.}

\subsubsection{Human Evaluation}

To further validate the effectiveness of our method, we conducted a further human study to evaluate the effectiveness of our method. Specifically, we randomly sample 100 instructions from the WizardLM test set to form a new instruction set. Then $3$ human participants are given the task of comparing the responses generated by the comparing models with the criteria same as the previous pair-wise evaluation, i.e. Helpfulness, Relevance, Accuracy, and Level of Detail. For each comparison, 3 options are given (Win, Tie, and Loss) and the final results are determined by the majority voting of the participants. The human studies are conducted based on LLaMA2-7B on two models: 
(1) Alpaca 5\% VS. Alpaca 100\%. 
(2) Alpaca-GPT4 5\% VS. Alpaca-GPT4 100\%.

\section{Experimental Result}

\begin{table*}[!htbh]
\centering
\scalebox{0.8}{
\begin{tabular}{ll|c|ccccc|c}
\toprule
\textbf{Dataset/} &\textbf{Superfilter} &\textbf{Pairwise $\uparrow$}& \multicolumn{5}{c|}{\textbf{Huggingface Open LLM Leaderboard} $\uparrow$} & \textbf{AlpacaEval $\uparrow$} \\
\textbf{Base Model} &\textbf{Ratio(Size)} & Winning Score & \textbf{Average} & ARC & HellaSwag & MMLU & TruthfulQA & Win Rate \\
\midrule
\textbf{Alpaca/} &100\% & 1.000 & 55.25 & 54.35 & 78.65 & 47.02 & 40.98 & 27.75 \\
\textbf{LLaMA2-7B} &5\%(2,600) & \textbf{1.133} & \textbf{55.6}7 & 56.57 & 80.15 & 45.21 & 40.74 & \textbf{33.04} \\
&10\%(5,200) & \textbf{1.101} & \textbf{56.97} & 58.02 & 80.57 & 47.16 & 42.14 & - \\
&15\%(7,800) & \textbf{1.193} & \textbf{56.61} & 56.23 & 80.29 & 46.73 & 43.21 & - \\
\midrule
\textbf{Alpaca/} &100\% & 1.000 & 58.78 & 57.59 & 81.98 & 54.05 & 41.49 & 35.00 \\
\textbf{LLaMA2-13B} &5\%(2,600) & \textbf{1.174} & \textbf{60.96} & 61.60 & 83.84 & 55.79 & 42.63 & \textbf{45.71} \\
&10\%(5,200) & \textbf{1.069} & \textbf{61.11} & 62.12 & 83.74 & 55.09 & 43.50 & - \\
&15\%(7,800) & \textbf{1.142} & \textbf{60.90} & 60.92 & 83.58 & 55.24 & 43.86 & - \\
\midrule
\textbf{Alpaca-GPT4/} &100\% & 1.000 & 58.71 & 54.69 & 80.05 & 47.89 & 52.21 & 71.32 \\
\textbf{LLaMA2-7B} &5\%(2,600) & \textbf{1.014} & \textbf{59.66} & 56.74 & 81.19 & 46.80 & 53.92 & \textbf{72.13} \\
&10\%(5,200) & \textbf{1.064} & \textbf{59.80} & 57.42 & 81.79 & 45.67 & 54.33 & - \\
&15\%(7,800) & \textbf{1.078} & \textbf{60.02} & 57.00 & 81.21 & 46.15 & 55.72 & - \\
\midrule
\textbf{Alpaca-GPT4/} &100\% & 1.000 & 60.81 & 57.94 & 82.22 & 54.84 & 48.25 & 77.86 \\
\textbf{LLaMA2-13B} &5\%(2,600) & \textbf{1.041} & \textbf{63.29} & 62.29 & 84.96 & 55.78 & 50.13 & \textbf{78.15} \\
&10\%(5,200) & \textbf{1.046} & \textbf{63.65} & 62.63 & 84.51 & 55.39 & 52.06 & - \\
&15\%(7,800) & \textbf{1.078} & \textbf{63.65} & 62.88 & 84.32 & 55.35 & 52.05 & - \\
\bottomrule
\end{tabular}
}
\caption{
\textbf{Comparison of Superfiltering with four data selection ratios} (5\%, 10\%, 15\%, 100\%) when finetuning two LLMs (LLaMA2-7B/13B) on two datasets (Alpaca and Alpaca-GPT4). The finetuned models are evaluated by the pair-wise winning score (comparison to the baseline model finetuned on 100\% data), Open LLM Leaderboard, and AlpacaEval. In the parathesis are the ratio of data being used and its exact number. The winning score is calculated as (Num(Win)$-$Num(Lose))$/$Num(All) $+ 1$, where the win-tie-lose numbers are reported in Figure \ref{pair}. The consistent improvement on all the \textbf{three} evaluation benchmarks demonstrates the effectiveness of Superfitering. 
}
\label{tbl:bench}
\end{table*}

\subsection{Main results}

In this section, we present the evaluation results of three different evaluation settings as described in the previous section as shown in Table \ref{tbl:bench}. The \textbf{Pair-Wise Winning Score} indicates the result directly comparing with the corresponding model trained with full data. These values that are greater than $1.0$ represent better responses generated by our Superfiltering models than full data models. The detailed win-tie-lose numbers are presented in Figure \ref{pair}. 
Moreover, the performance of our models and baseline models on the \textbf{Huggingface Open LLM Leaderboard } and the \textbf{AlpacaEval Leaderboard} are also presented in Table \ref{tbl:bench} where we can see our models using $5\%$, $10\%$, $15\%$ data outperform the models trained with full data on both benchmarks on both LLaMA2-7B and LLaMA-13B settings. These results further showcase the effectiveness of our Superfiltering. Moreover, the usefulness of Superfiltering on the high-quality Alpaca-GPT4 dataset further shows the potential of our method, which is astonishing that \textit{a weak language model like GPT-2 is able to filter the data responses generated by GPT-4. }

For the \textbf{human evaluation}, we compare the performances between models trained with 5\% Superfiltring data and full data based on LLaMA2-7B models, on Alpaca and Alpaca-GPT4 Datasets. In the comparison (1) Alpaca 5\% VS. Alpaca 100\%, our model wins on $50$ out of $100$ instruction, ties on $18$, and losses on $32$ instructions. In the comparison (2) Alpaca-GPT4 5\% VS. Alpaca-GPT4 100\%, our model wins on $49$ out of $100$ instruction, ties on $5$, and losses on $46$ instructions. This human evaluation of Superfiltering further validates our method.


\begin{table}[!t]
\centering
\scalebox{0.85}{
\begin{tabular}{llcccc}
\toprule
\multicolumn{2}{l}{Ablation} & \multicolumn{3}{c}{Pairwise Winning Score $\uparrow$}\\
\midrule
\multicolumn{2}{l}{Data Selection Budget} & \textbf{5\%} & \textbf{10\%} & \textbf{15\%}  \\
\midrule
Strategy: &Random  & 0.936     & 0.968       & 0.977     \\
&Diversity  & 0.927     & 0.977       & 0.982     \\
&Perplexity  & 0.261     & 0.569       & 0.610     \\
\midrule
Filter: &GPT-2-large  & 1.165  & 1.046  & 1.193  \\
&GPT-2-XL  & 1.064& 1.165  & 1.128  \\
&GPT-NEO  & 1.096 & 1.197 & 1.156    \\
&LLaMA2-7B  & 1.303 & 1.330 & 1.294\\
\midrule
\multicolumn{2}{l}{Superfilter (IFD, GPT-2)} & 1.133     & 1.101       & 1.193     \\
\bottomrule
\end{tabular}
}
\caption{
\textbf{Ablation study of data selection strategies and filter models} on finetuning LLaMA2-7B using the Alpaca dataset. The pairwise winning score compares each finetuned model with the full-data finetuned model and computes (Num(Win)$-$Num(Lose))$/$Num(All) $+ 1$. All the comparisons are performed by GPT-4 on the WizardLM test set.
}
\label{tbl:ablation_scores}
\end{table}

\subsection{Comparison with Other Methods }

In this subsection, we compare our method with three other widely accepted instruction-tuning data selection methods on LLaMA2-7B using the Alpaca Dataset, in terms of performance and efficiency.  
``ChatGPT score'' represents utilizing ChatGPT to evaluate the quality of the Alpaca data samples proposed by \citet{chen2023alpagasus}. 
``Reward-model score'' represents utilizing extra reward models to rate the given data samples, in this experiment, ``OpenAssistant/reward-model-deberta-v3-large-v2'' is utilized following \citet{bukharin2023data, du2023mods}. 
``IFD score'' represents directly utilizing the base model to be trained to calculate the Instruction-Following Difficulty scores proposed by \citet{cherry}. 

\begin{table}[!t]
\centering
\scalebox{0.72}{
\begin{tabular}{l|ccc|c}
\toprule
\multicolumn{1}{l|}{Comparison} & \multicolumn{3}{c|}{Pairwise Winning Score $\uparrow$} & Time\\
\midrule
\multicolumn{1}{l|}{Data Selection Budget} & \textbf{5$\%$} & \textbf{10$\%$} & \textbf{15$\%$} & (min) \\
\midrule
Superfiltering (ours)  & -     & -       & - & 8      \\
vs. ChatGPT score  & 1.028     & 1.174       & 1.170  & 120   \\
vs. Reward score  & 1.280     & 1.096       & 1.147   & 1400  \\
vs. IFD score (LLaMA2-7B)  & 0.853     & 0.761       & 0.927 & 161    \\
\bottomrule
\end{tabular}
}
\caption{\textbf{Comparison with Other Methods} in terms of performance and efficiency. The Pairwise Winning Scores are calculated between models using our method and other methods. All the comparisons are performed by GPT-4 on the WizardLM test set and the values that are greater than $1.0$ represent our models are better and vice versa. The time shown represents the time used for data filtering. \looseness-1
}
\label{tbl:compare_others}
\end{table}

As shown in the Table \ref{tbl:compare_others}. Superfiltering outperforms ChatGPT and Reward-model scores in terms of winning scores and filtering efficiency. 
Compared with Superfiltering, utilizing ChatGPT for selection needs further costs on utilizing API models, and its efficiency is largely constrained by the rate limits set by the API company. 
Utilizing outside reward models is a cost-free method, while the efficiency can not be guaranteed: it is $175\times$ slower than our Superfiltering. 
Directly calculating IFD scores based on the base model is a promising method as it is the only method that has a better performance than our Superfiltering, as it selects the data that best matches the model. 
However, our Superfilrtering is $20\times$ faster. 
From the comparison, it is observed that our model largely reduces the time used for data filtering, and we are the only method that makes the filtering time shorter than the later training time, making this process feasible in practical usage. 
Moreover, our filtering method can be implemented on consumer-level GPUs with much smaller graphic memories like 6 GB since we only utilize the GPT-2 model, while all other methods require large industry-level GPUs.

\subsection{Ablation Study}

In this subsection, extensive ablation experiments are conducted to validate the effectiveness of our Superfiltering. The experiments are performed on the LLaMA2-7B model using the Alpaca dataset. Our focus is on two aspects: the impact of different data selection strategies and the effect of using various language models for data selection. All models are trained under the same settings.

As shown in Table \ref{tbl:ablation_scores}, in addition to our method ``Superfiltering (GPT-2)'', we also try several baseline strategies: ``Random'' represents the models trained with randomly selected data. ``Diversity'' represents the models trained with data considering only diversity, by utilizing the k-means algorithm. ``Perplexity'' represents the models trained with data based on the perplexity calculated on GPT-2. Moreover, the lower part of the table lists the models using the IFD score to select the training subset, powered by other language models. The performances of models are assessed by the pair-wise winning score, which is calculated as (Num(Win)$-$Num(Lose))$/$Num(All) $+ 1$, and all the comparisons are performed by GPT4 on the WizardLM test set.
\looseness-1

As shown in Table \ref{tbl:ablation_scores}, compared with other strategies, models trained with our method consistently outperform the models trained on the full dataset, indicating the efficacy of our method. Regarding the impact of different language models, whichever language model is utilized to calculate the IFD scores, the corresponding models would surpass the baseline model, indicating the strong consistency and transferability of the IFD score as the selection metric. Moreover, the models using LLaMA2-7B reasonably achieve the highest performance, due to the consistency between the model to calculate the IFD scores and the model to be trained.

\section{Further Discussion}



\subsection{``Plug-and-Play'' without Additional Training}

Our Superfiltering introduces a transformative advantage: the unnecessity of training for even weak language models and the unnecessity of the extra hold-out sets. 

Traditional proxy-based methods like \citet{Coleman2020Selection} and \citet{nguyen-etal-2022-famie} are required to further train weak models to bridge the performance gap with stronger models.
In the context of instruction tuning data selection, model training or a hold-out set is always necessary if no extra strong models like ChatGPT or other trained reward models are utilized. 
\citet{lu2023instag} utilizes chatGPT to tag the instruction datasets and train LLMs for tagging instruction samples based on these data. 
\citet{li2023shot} requires a holdout test set as the indicator for the data improvement. 
\citet{cao2023instruction} proposes to train several models based on losses on unseen datasets and obtain a regression model to estimate data qualities. \looseness-1

However, our study reveals that pre-trained weak models are naturally effectively capable of acting as proxies for strong models when utilizing the IFD for data selection.
These models do not necessitate fine-tuning on specific datasets, thereby reducing the risk of out-of-distribution issues. 
This innovative approach not only simplifies the data selection process but also revolutionizes the efficiency and applicability of such methods in large language model instruction tuning.




\subsection{Sperfiltering as Dataset Assessment}

Our finding, the consistency of perplexity-based metrics across weak and strong language models, might push forward the usage of small language models as proxy models, not only in the area of data selection. Moreover, our finding provides an efficient and general way to \textbf{assess the instruction data of the whole dataset.} 

As illustrated in Figure \ref{ifd_dis}, a preliminary assessment of the dataset can be made by examining the violin plots of the Instruction Followed Difficulty (IFD) scores. 
The classic Alpaca dataset, known for its relatively lower quality, exhibits a violin plot with wide upper and lower sections, indicating significant variance in data quality. The WizardLM dataset, generated by prompting ChatGPT 3.5, displays complex instructions but is affected by generation noise, reflected by its violin plot which is wide at the top and has a long, narrow tail. In contrast, the Alpaca-GPT4 dataset, created using GPT-4, demonstrates a higher quality lower bound, as evidenced by its violin plot with a more compressed lower section, indicating fewer low-quality instructions. 
To our knowledge, there is no visualization that can represent the quality of a dataset with different characteristics, (some work might draw the distribution of verb-noun pairs, but it only represents the diversity and can not reflect the overall quality of the dataset). 

\subsection{Why Weak-to-Strong Consistent?}

We hypothesize this Weak-to-Strong consistency can be explained from (1) the \textbf{perspective of Perplexity} and (2) the \textbf{perspective of String Pattern}, presented in Appendix \ref{why}.

\subsection{Superfiltering with Diversity}

We further incorporate the diversity metric into the Superfiltering, which further reduces the data used for training, presented in the Appenix \ref{SuperfilteringD}.

\section{Related Work}


\subsection{Instruction Tuning Data Selection}

{Instruction tuning \cite{wei2022finetuned, sanh2022multitask, Longpre2023TheFC, liu2023mmc} is a widely-used training paradigm to equip LLMs with the instruction-following ability. 
To further select the data for more efficient instruction tuning, existing automatic data selection methods mainly utilize extra LLMs for the selection. 
\citet{lu2023instag} utilizes proprietary chatGPT to tag the instruction data to ensure diversity and complexity. 
\citet{chen2023alpagasus} utilizes proprietary LLMs chatGPT and Claude2 to assess the quality of the instruction data, generating both ratings and explanations. 
\citet{du2023mods} and \citet{bukharin2023data} utilize an extra reward model to assess the quality of data and utilize these scores as a part of their method. 
\citet{cherry} firstly proposes a self-guided method in which no extra LLMs are utilized but still needs to calculate Instruction-Following Difficulty (IFD) scores based on the original pre-trained LLM. 
Though effective, these methods overly rely on large language models and are too time-consuming to put into practical use.

\subsection{Small Model Proxies for Large Models}
The use of proxy models is increasingly recognized in machine learning, particularly when resources are constrained or there is a limited understanding of the original model's architecture. \citet{chen-etal-2023-rev} and \citet{hase-etal-2020-leakage} demonstrate the utility of lightweight proxy models in evaluating free-text rationales. Similarly, \citet{puigcerver2021scalable} leverages embeddings from expert models with a k-nearest neighbors classifier to simplify the training of more complex systems. \citet{Coleman2020Selection} and the FAMIE \cite{nguyen-etal-2022-famie} apply downscaled proxy models in fields like image classification and information extraction, utilizing techniques such as layer removal and knowledge distillation for aligning these proxies with larger models. 
\citet{burns2023weaktostrong} explores the concept of enhancing larger models through weak supervision, and training on weaker model labels. 

\section{Conclusion}

In this paper, we reveal the consistency between weak and strong language models in perceiving instruction difficulties. Based on the consistency, we present ``Superfiltering'', a novel and efficient approach for data filtering in the instruction tuning of LLMs. By effectively utilizing weaker models as proxies for evaluating instructional data, we achieve a significant leap in efficiency, accelerating the data filtering process largely. The experimental results affirm that our method considerably reduces computational overhead while maintaining or even improving the instruction tuning performance of LLMs. It is astonishing that a weak language model like GPT-2 is able to filter the data responses generated by GPT-4. 
Our Superfiltering marks a substantial contribution by offering a scalable, resource-efficient, and effective strategy for the advancement of AI technologies.

\section*{Limitations}


Our study introduces ``Superfiltering'' as an innovative approach to enhance the instruction tuning process of LLMs through efficient data selection using smaller models. While we have observed promising results, several areas warrant further exploration:
(1) Superfiltering primarily utilizes IFD scores to select data, focusing on instruction difficulty. Future enhancements should include additional dimensions such as data diversity. 
(2) There are $3$ datasets and $2$ LLaMA2 base models are involved, 
further research is recommended to explore its effectiveness with a broader array of LLMs and Datasets.

\bibliography{custom}

\appendix



\clearpage
\section{Superfiltering with Diversity}
\label{SuperfilteringD}

Though Superfiltering shows promising performance with great efficiency, it can also be implemented together with other dimensions such as data diversity. 
Motivated by recent work that further includes Diversity metrics in the data selection process, we introduce an extended version of Superfiltering, Superfiltering with Diversity (\textbf{Superfiltering.D}). 

We hypothesize that the diversity metrics work better when implemented on a high-quality data subset than the whole dataset with mixed quality. Thus we propose to first utilize Superfiltering to select a subset with relatively high quality, then further utilize Facility Location Function \footnote{\url{https://apricot-select.readthedocs.io/en/latest/functions/facilityLocation.html}} to further compress the selected data. Compared with other diversity metrics, the Facility Location Function can strike a balance between capturing diversity and ensuring the representation of different clusters or regions within the data, it ensures a global view of the given high-quality subset. 

To further preserve the efficiency of our Superfiltering.D, we utilize ``sentence-transformers/all-MiniLM-L6-v2'' \cite{reimers-gurevych-2019-sentence} as the encoder, which only has approximately 80M parameters. In our preliminary experiments on the Alpaca and Alpaca-GPT4 dataset, where we first select $20\%$ of the data by Superfiltering, then utilize the Facility Location Function to further select $2\%$ of the data. The models trained with $2\%$ of the data have a comparable or better performance than full data models.

\begin{table*}[!htbh]
\centering
\scalebox{0.8}{
\begin{tabular}{ll|c|ccccc}
\toprule
\textbf{Dataset/} &\textbf{Superfiltering} &\textbf{Pairwise $\uparrow$}& \multicolumn{5}{c}{\textbf{Huggingface Open LLM Leaderboard} $\uparrow$}  \\
\textbf{Base Model} &\textbf{Ratio(Size)} & Winning Score & \textbf{Average} & ARC & HellaSwag & MMLU & TruthfulQA  \\
\midrule
\textbf{Alpaca/} &100\% & 1.000 & 55.25 & 54.35 & 78.65 & 47.02 & 40.98  \\
\textbf{LLaMA2-7B} &2\%(1,040) & \textbf{1.028} & \textbf{55.43} & 55.97 & 79.89 & 45.51 & 40.34 \\
\midrule
\textbf{Alpaca-GPT4/} &100\% & 1.000 & \textbf{58.71} & 54.69 & 80.05 & 47.89 & 52.21 \\
\textbf{LLaMA2-7B} &2\%(1,040) & \textbf{1.078} & 58.70 & 56.48 & 80.30 & 45.23 & 52.79 \\
\midrule
\end{tabular}
}
\caption{
Comparison between Superfiltering.D with 2\% data and full data model on two datasets (Alpaca and Alpaca-GPT4). The finetuned models are evaluated by the pair-wise winning score (comparison to the baseline model finetuned on 100\% data) and Open LLM Leaderboard. In the parathesis are the ratio of data being used and its exact number. 
}
\label{tbl:bench}
\end{table*}

\clearpage
\section{Evaluation Metric}
\label{evaluation}

\subsection{Pair-wise comparison}

Evaluating responses generated by Large Language Models (LLMs) like GPT-4 remains a complex and ongoing research area, particularly for open-domain questions where establishing a clear ground truth is challenging. Traditional methods often fall short in assessing the instruction-following ability of these models. Recent trends, however, involve using LLMs themselves, such as GPT-4, as evaluators, a practice that has gained widespread acceptance in the field \cite{touvron2023llama2, vicuna2023, dettmers2023qlora, liu2023geval}. Previous studies \cite{zheng2023judging, alpaca_eval, sottana-etal-2023-evaluation} have shown that GPT4's evaluations are consistent with human evaluations. We utilized the testing instruction set from WizardLM \cite{xu2023wizardlm} and Vicuna \cite{vicuna2023} which contain $218$ and $80$ diverse human-curated instructions respectively.

Our study adopts the evaluation strategy as outlined by \citet{chen2023alpagasus, cherry, Li2023ReflectionTuningDR}, involving a detailed rating system for model-generated responses. Each response is scored reflecting various dimensions such as the accuracy and relevance of the response. This method is in line with previous research efforts to assess the effectiveness of language models more accurately. 
Moreover, to address the issue of positional bias, as discussed in the works of \citet{ko-etal-2020-look, wang2023large}, we present the responses generated by the model in two separate sequences for evaluation by the LLM judge. This approach aims to ensure a more balanced and unbiased assessment of the model's performance. Then for each instruction, we compare the responses by "Win-Tie-Loss".

\subsection{AlapcaEval Leaderboard}

The AlpacaEval Leaderboard, utilizing the AlpacaFarm \cite{dubois2023alpacafarm, alpaca_eval} evaluation dataset, is an automated, efficient, and reliable evaluation tool for LLMs. It benchmarks LLMs' performance in following generic user instructions by comparing their outputs with those from Davinci003, demonstrating high alignment with human expert annotations. AlpacaFarm, underlying AlpacaEval, is a cost-effective simulator for research on learning from human feedback, significantly reducing the time and cost traditionally associated with such studies. While AlpacaEval offers valuable insights, it primarily focuses on simpler instructions and does not encompass safety evaluations or complex tasks, and its evaluation may correlate win rates with response lengths. These tools represent significant advancements in LLM evaluation and development, enabling more accessible and diverse research. Considering our budget, we only run the evaluation on $5\%$ settings.  

\subsection{Open LLM Leaderboard}

The Huggingface Open LLM Leaderboard, incorporating the evaluation method from the Eval Harness \cite{eval-harness}, serves as a comprehensive framework for evaluating generative language model capabilities. It focuses on four critical benchmarks: ARC \cite{clark2018think}, HellaSwag \cite{zellers-etal-2019-hellaswag}, MMLU \cite{hendrycks2021measuring}, and TruthfulQA \cite{lin-etal-2022-truthfulqa}. These benchmarks test the models on various aspects, such as reasoning, common-sense understanding, and factual accuracy. The leaderboard offers an effective platform for comparing different LLMs, providing valuable insights into their performance across these diverse and challenging tasks





\clearpage
\section{Prompt for Evaluation}
\label{p_evaluation}

The detailed pair-wise comparison prompt for the pair-wise comparison is in Figure \ref{appendix_prompt}. 

\begin{figure}[h]
  \centering
  \parbox{0.48\textwidth}{
        \rule{0.48\textwidth}{1.5pt} 
        Prompt for Performance Evaluation \\
        \rule{0.48\textwidth}{0.8pt} 
        \textbf{System Prompt} \\
        You are a helpful and precise assistant for checking the quality of the answer. \\

        \textbf{User Prompt} \\
        \text{[Question]}\\
        \textit{Question}\\
        \text{[The Start of Assistant 2's Answer]}\\
        \textit{Answer 2}\\
        \text{[The End of Assistant 2's Answer]}\\
        \text{[The Start of Assistant 2's Answer]}\\
        \textit{Answer 2}\\
        \text{[The End of Assistant 2's Answer]}\\

        We would like to request your feedback on the performance of two AI assistants in response to the user question displayed above. \\
        Please rate the helpfulness, relevance, accuracy, level of details of their responses. Each assistant receives an overall score on a scale of 1 to 10, where a higher score indicates better overall performance. \\
        Please first output a single line containing only two values indicating the scores for Assistant 1 and 2, respectively. The two scores are separated by a space. In the subsequent line, please provide a comprehensive explanation of your evaluation, avoiding any potential bias and ensuring that the order in which the responses were presented does not affect your judgment.

        \rule{0.48\textwidth}{0.8pt} 

  }
\caption{
The prompt we used to request ChatGPT or GPT4 to evaluate the responses. 
} 
\label{appendix_prompt} 
\end{figure}

\clearpage
\section{Why Weak-to-Strong Consistent?}
\label{why}

In this section, we provide the hypothesis on why this Weak-to-Strong consistency exists on two perspectives. 

\textbf{From the perspective of Perplexity}: The consistency between IFD scores is built on the consistency of perplexities since it is the ratio between two kinds of perplexities. As for the reason why the rankings of perplexities are consistent between small and large language models, we hypothesize that both small and large language models are trained on similar corpus (almost all the existing data corpus), which means they're exposed to the same underlying distribution of language. Even though larger models can capture more nuances and details due to their increased parameter count, both sizes of models are ultimately trying to approximate the same underlying language structure. 

\textbf{From the perspective of String Pattern}: We also examined verb-noun pairs of the data samples filtered by IFD scores on both weak (GPT-2) and strong (LLaMA2-7B) language models on the Alapca dataset and found a strong consistency in these verb-noun pairs. The results are listed in Table \ref{tbl:vn}. 
From the table, we can find out that the language models tend to assign \textbf{higher IFD scores to those data that require a lot of creativity, thinking skills, and deep understanding} while assigning \textbf{lower IFD scores to those data that are more about following rules and need less creativity.} This holds true for both small and large language models. 

\begin{table*}[h!]
\centering
\scalebox{1}{
\begin{tabular}{llc|llc}
\toprule
\multicolumn{3}{c|}{\textbf{LLaMA2-7B (Top 5\% IFD)}}  & \multicolumn{3}{c}{\textbf{GPT2 (Top 5\% IFD)}}  \\
\midrule
\textbf{Verb} & \textbf{Noun} & \textbf{Count} & \textbf{Verb} & \textbf{Noun} & \textbf{Count} \\
\midrule
write & story & 119 & write & story & 117 \\
generate & list & 104 & generate & story & 84 \\
generate & story & 89 & create & story & 52 \\
write & essay & 61 & generate & list & 47 \\
create & list & 44 & write & essay & 39 \\
write & post & 41 & create & list & 29 \\
write & article & 39 & write & paragraph & 26 \\
create & story & 36 & write & post & 25 \\
generate & recipe & 34 & generate & recipe & 19 \\
create & recipe & 31 & give & example & 17 \\
\midrule
\multicolumn{3}{c|}{\textbf{LLaMA2-7B (Least 5\% IFD)}}  & \multicolumn{3}{c}{\textbf{GPT2 (Least 5\% IFD)}} \\
\midrule
\textbf{Verb} & \textbf{Noun} & \textbf{Count} & \textbf{Verb} & \textbf{Noun} & \textbf{Count} \\
\midrule
rewrite & sentence & 140 & rewrite & sentence & 124 \\
edit & sentence & 56 & edit & sentence & 74 \\
classify & sentence & 51 & classify & sentence & 30 \\
change & sentence & 42 & replace & word & 24 \\
convert & sentence & 33 & change & sentence & 23 \\
arrange & word & 29 & construct & query & 20 \\
rearrange & word & 27 & convert & sentence & 20 \\
categorize & sentence & 26 & arrange & word & 17 \\
find & word & 24 & find & word & 17 \\
classify & text & 18 & classify & animal & 16 \\
\bottomrule
\end{tabular}
}
\caption{Verb-noun pairs of the top and least 5\% IFD scores calculated by LLaMA2-7B and GPT2 on the Alpaca dataset.}
\label{tbl:vn}
\end{table*}

\end{document}